\documentclass[letterpaper, 10 pt, conference]{ieeeconf}  

\IEEEoverridecommandlockouts                              

\usepackage{graphics} 
\usepackage{epsfig} 
\usepackage{amsmath} 
\usepackage{subcaption}
\usepackage{booktabs}
\usepackage{multirow}

\title{\LARGE \bf
Physics-Informed Real NVP for Satellite Power System Fault Detection
}

\author{Carlo Cena$^{1}$, Umberto Albertin$^{1}$, Mauro Martini$^{1}$, Silvia Bucci$^{2}$ and Marcello Chiaberge$^{1}$
\thanks{*This publication has received funding from the MUR – DM 351/2022 and is part of the project PNRR-NGEU which has received funding from the MUR – DM 117/2023.}
\thanks{$^{1}$Carlo Cena, Umberto Albertin, Mauro Martini and Marcello Chiaberge are with the Department of Electronics and Telecommunications (DET),
        Politecnico di Torino, Corso Duca degli Abruzzi, 24, 10129 Torino (TO), Italy
        {\tt\small name.surname@polito.it}}%
\thanks{$^{2}$Silvia Bucci is with Argotec SRL,
        via Luigi Burgo, 8, 10099 San Mauro Torinese (TO), Italy
        {\tt\small silvia.bucci@argotecgroup.com}}%
}
\begin{document}

\maketitle
\thispagestyle{empty}
\pagestyle{empty}

\begin{abstract}
The unique challenges posed by the space environment, characterized by extreme conditions and limited accessibility, raise the need for robust and reliable techniques to identify and prevent satellite faults. Fault detection methods in the space sector are required to ensure mission success and to protect valuable assets. In this context, this paper proposes an Artificial Intelligence (AI) based fault detection methodology and evaluates its performance on ADAPT (Advanced Diagnostics and Prognostics Testbed), an Electrical Power System (EPS) dataset, crafted in laboratory by NASA.

Our study focuses on the application of a physics-informed (PI) real-valued non-volume preserving (Real NVP) model for fault detection in space systems. The efficacy of this method is systematically compared against other AI approaches such as Gated Recurrent Unit (GRU) and Autoencoder-based techniques.

Results show that our physics-informed approach outperforms existing methods of fault detection, demonstrating its suitability for addressing the unique challenges of satellite EPS sub-system faults. Furthermore, we unveil the competitive advantage of physics-informed loss in AI models to address specific space needs, namely robustness, reliability, and power constraints, crucial for space exploration and satellite missions.
\end{abstract}

\section{INTRODUCTION}

Satellites are essential for many applications such as communication \cite{rahmat2014technology}, navigation \cite{sat_navigation}, and remote sensing. Earth observation can provide substantial support to monitor climate changes \cite{rem_sensing_1}, model and predict flood and environmental disasters \cite{rem_sensing_2}, and extract precious vegetative trends for agriculture \cite{martini2021domain}. Besides that, space exploration is a rapidly growing economic sector that is pushing the evolution of cutting-edge technology in the satellite field. For example, small size CubeSats have been recently employed for asteroid impact observation \cite{deep_space_1} and deep space biological exploration \cite{deep_space_2}. However, operating in the harsh and unpredictable space environment exposes them to various sources of faults that can compromise their functionality and performance \cite{faults_type}.

Traditional fault detection methods based on rule-based \cite{schein2006rule} or model-based \cite{isermann2005model} techniques have limitations in handling complex and uncertain scenarios, especially when dealing with high-dimensional and nonlinear data.

\begin{figure}[t]
    \centerline{\includegraphics[width=\linewidth]{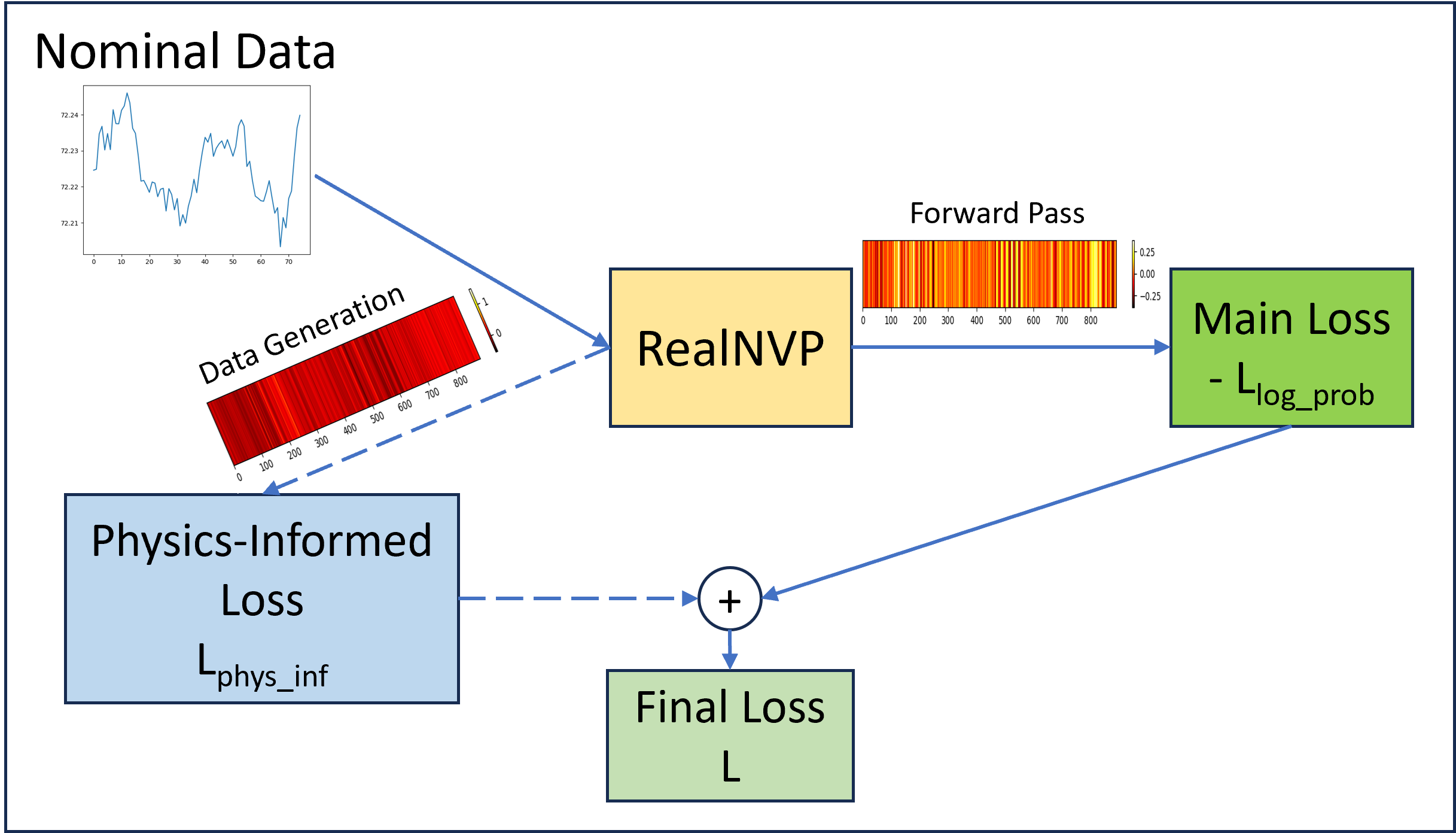}}
    \caption{Training step for the proposed fault detection pipeline with Real NVP. Nominal data is fed into the model and a loss composed of two terms, \(-L_{log\_prob}\) and \(L_{phys\_inf}\), is evaluated. The former, which is the standard loss used with these models, is computed starting from the output of the forward pass. The latter is a physics-informed loss that boosts the performance and interpretability of the model, generated by the inverse propagation of Real NVP.}
    \label{fig:training_method_realnvp}
\end{figure}

Therefore, there is a growing interest in applying Artificial Intelligence (AI) methods to enhance the fault detection capabilities of satellites \cite{ai_fd_generic}, \cite{framework_shallow_space}. AI methods can learn from data and extract useful features for fault detection, without relying on prior knowledge or assumptions. Many works tried to exploit time series data patterns to learn fault detection and diagnostic \cite{lstm_military_sat}, \cite{lstm_graphnn_data_col_normed}, \cite{time_series_realnvp}, \cite{bayesian_rnn}, \cite{gru_space}. However, most of the existing AI methods for fault detection in the space sector are based on black-box models that do not incorporate any physical information or constraints. This can lead to unrealistic or inaccurate results, especially when dealing with rare or novel faults. Moreover, black-box models are often computationally expensive and require large amounts of data, which are not always available or feasible in the space domain \cite{ai_diff_datasets_sat}.

In this paper, we propose a novel AI method for fault detection in satellite power systems, based on a physics-informed (PI) \cite{pinn} real-valued non-volume preserving (Real NVP) model. Real NVP \cite{realnvp_paper} is a type of generative model that can learn the probability distribution of the data and generate realistic samples. By incorporating physical information into the loss function of Real NVP, we can enforce the physical consistency and plausibility of the generated samples, as well as reduce the data and computational requirements. We apply our PI-Real NVP model to the ADAPT dataset \cite{adapt_paper}, an Electrical Power System (EPS) dataset created by NASA in a laboratory setting to simulate various types of faults. We compare the performance of our model with other AI-based methods such as Gated Recurrent Unit (GRU) \cite{gru_paper} and Autoencoder-based techniques \cite{autoenc_paper, autoenc_virtual_adapt}, and with previous approaches on ADAPT such as the convolutional-based approach proposed in \cite{cnn_adapt} and the Multilayer Perceptron (MLP) adopted in \cite{mlp_adapt}. The fault detection models have been trained with a semi-supervised setup \cite{semi_supervised}: the training follows the unsupervised paradigm of the Real NVP to learn the nominal data distribution, whilst faulty data are filtered out with human supervision to identify the nominal training data. Faulty data are then used to evaluate the models during testing and validation. We show that our model achieves superior results in terms of accuracy, robustness, and interpretability. We also demonstrate the advantages of using physics-informed loss in AI models for fault detection in space systems, such as reducing the false alarm rate, improving the generalization ability on new types of faults, and saving power consumption. Our work provides a new perspective on how to leverage the synergy between physics and AI for fault detection in satellites and other complex systems.

The contributions of this research are the following:
\begin{itemize}
    \item We tackle a satellite EPS fault detection problem with a novel semi-supervised Machine Learning approach;
    \item We introduced a physics-informed loss focused on voltage and current signal relationships of the given EPS sub-system and provided insights into the feature maps produced by the AI models;
    \item We demonstrate a competitive advantage on the ADAPT dataset outperforming other state-of-the-art methods.
\end{itemize}

\section{Methodology}
Here we delve into the training and test phases of each model and describe the losses used, paying particular attention to the physics-informed loss.

We focused on the application of a physics-informed - Real NVP \cite{realnvp_paper} model for fault detection within the ADAPT dataset\footnote{https://data.nasa.gov/dataset/ADAPT-Dataset/6gjh-n6gb/about\_data}. In this study, a Real NVP with a multivariate normal distribution is adapted to address the challenges associated with fault detection in the space sector, leveraging its inherent capacity to capture intricate relationships within the data.

Real NVP \cite{realnvp_paper} is a methodology employed in density estimation tasks, particularly within the domain of generative modeling. Developed as an extension of the Normalizing Flow \cite{norm_flow_1, norm_flow_2} framework, it provides a modeling approach, with tractable likelihood evaluation and efficient sampling, complex high-dimensional distributions by learning an invertible mapping between the data and a simpler latent space.

\begin{figure}[t]
    \centerline{\includegraphics[width=1.0\linewidth]{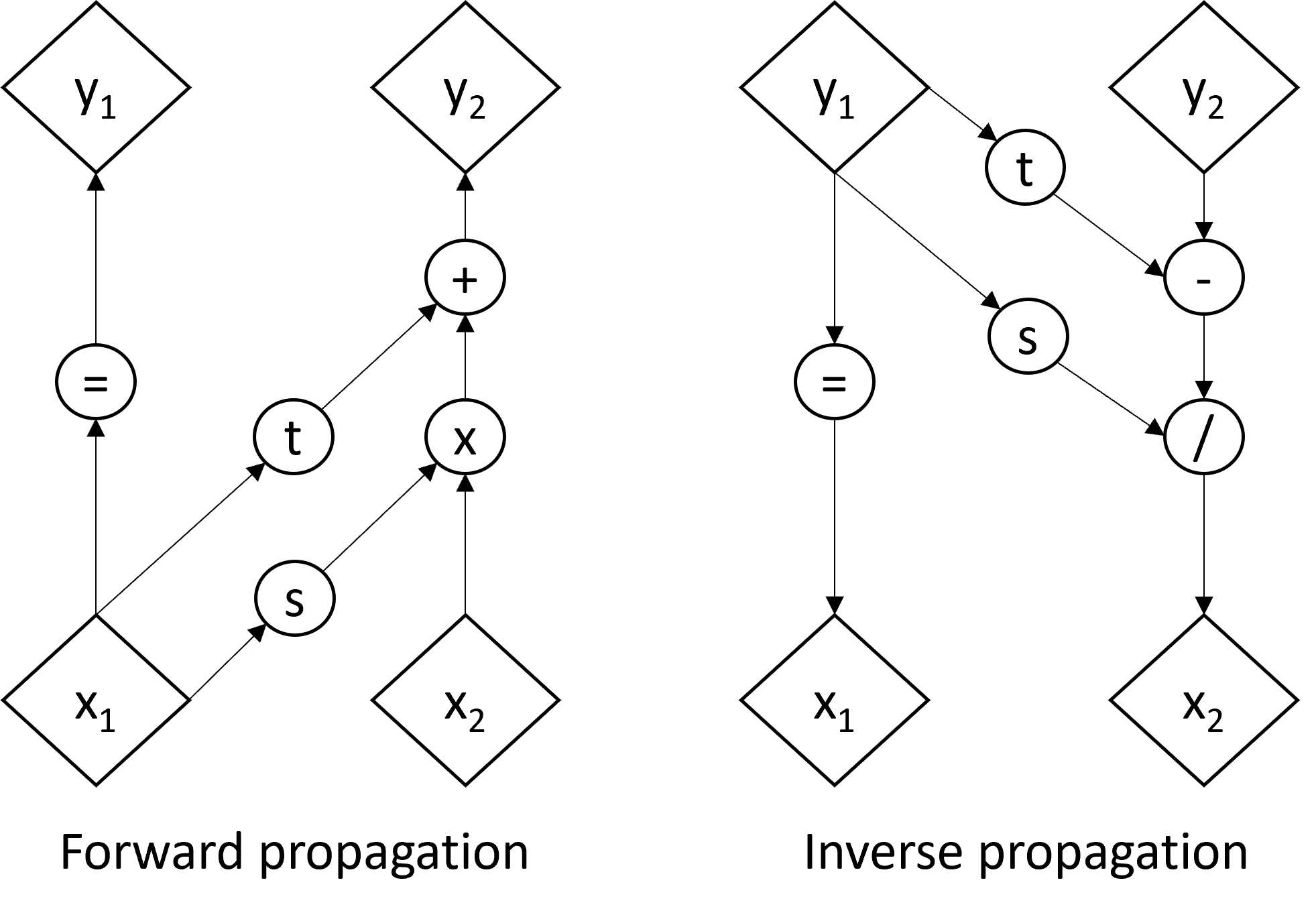}}
    \caption{Computational graph for propagation in both directions in a Coupling Layer \cite{realnvp_paper}. In this kind of model, \textit{s} is related to the scaling function while \textit{t} is related to the translation one. These functions assume the shape of neural networks with several layers that must be designed properly as a function of the application.}
    \label{fig:coup_layer}
\end{figure}

The key innovation of Real NVP with respect to other Normalising Flow models lies in the use of affine coupling layers (Fig. \ref{fig:coup_layer}), that can exploit the local dependencies within the data and capture diverse modes present in the data distribution, enabling the generation of realistic and diverse samples.

Each coupling layer operates by decomposing the data into two sets of variables and transforming one set while leaving the other unchanged. This transformation is achieved through a series of invertible, bijective functions, designed such that both forward and inverse mappings are computationally efficient, allowing for tractable and exact inference and generation of samples. These functions are typically implemented as neural networks.

Real NVP offers a scalable solution to model multi-modal distributions effectively for density estimation, making it suitable for large-scale datasets.

\subsection{Architectures}
Other than the Real NVP model, we implemented a set of other AI models to compare its performances. We analyzed the fault detection literature and selected a couple of the most used approaches: GRU \cite{gru_paper} and Autoencoders \cite{autoenc_paper}. GRU was selected as an alternative to the more used Long Short-Term Memory (LSTM) \cite{lstm_paper}, because it has been shown to lead to similar \cite{gru_similar} or better \cite{gru_better} performances with respect to the LSTM while being more computationally efficient.

\textbf{Autoencoder:} The Autoencoder \cite{autoenc_paper} used for the test is under-complete. More in detail, the hyper-parameters searched led to an encoder composed of two layers with 256 and 64 neurons, respectively. The decoder comprises two other layers with 256 neurons for the first and as many neurons as the number of input features for the second. Each layer is fully connected with a \textit{relu} activation function. The under-complete Autoencoder reduces the number of input features in a more dense configuration to capture the most salient information regarding the input data.

\textbf{GRU:} \cite{gru_paper} The best neural network architecture comprises two GRU layers. Each one has 256 neurons and the head is a fully connected layer with as many neurons as 50 times the input features of the ADAPT dataset, i.e. the time window considered by the model, to reconstruct the input data.

\textbf{Real NVP:} Real NVP \cite{realnvp_paper}, whose training procedure is described in the previous section, is a normalizing flow model. In this model, the neural network to be designed regards the scale \textit{s} and translation \textit{t} functions. \textit{t} is designed with four fully connected layers with 128 neurons and a fully connected layer with a number of neurons equal to the input features. The \textit{s} function is designed with the same structure as the \textit{t} transformation with a \textit{tanh} activation function for the last layer instead of the \textit{linear} activation function of the \textit{t}.

\begin{figure}[t]
\centerline{\includegraphics[width=1.0\linewidth]{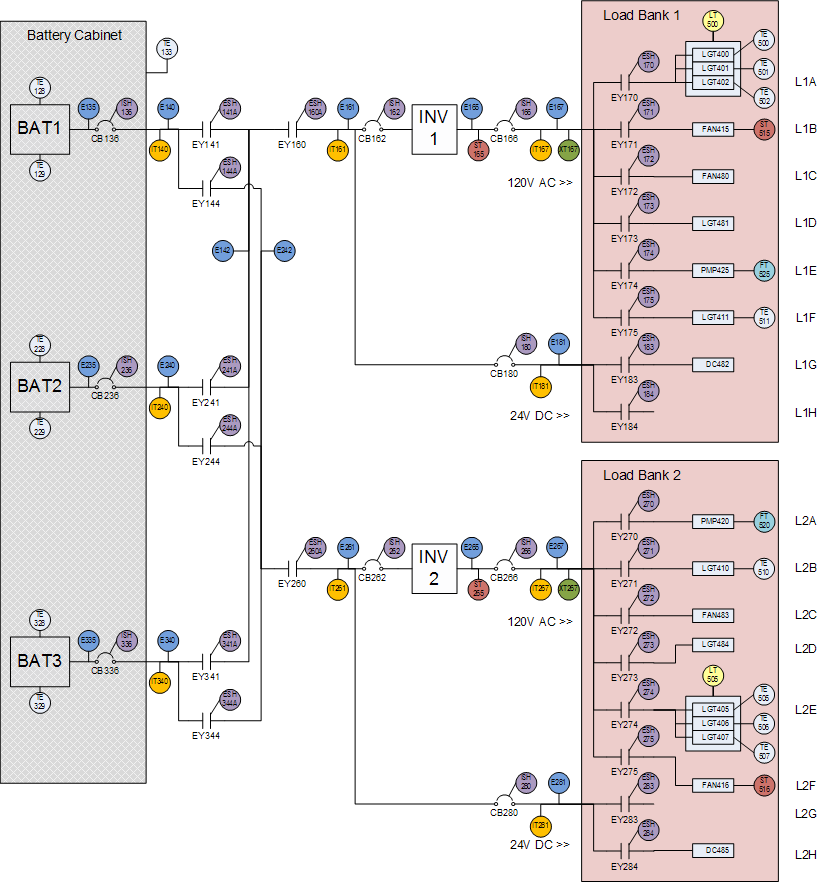}}
    \caption{Circuit of the testbed used to generate the ADAPT dataset. This circuit is used to handle the constraints related to the physics-informed loss composition.}
    \label{fig:circuit_adapt}
\end{figure}

\subsection{Loss Functions}
Moreover, we designed a physics-informed loss, to be used with all models, extracting physical relations from the circuit of the testbed assembled by NASA to create ADAPT, shown in Fig. \ref{fig:circuit_adapt}. This loss is composed of the following terms:

$$ d_{squared} =
    \frac{1}{|\boldsymbol{O}|}\sum_{i \in |\boldsymbol{O}|}\begin{cases}
     (x_{i} - 1)^2, & \text{if}\ x_{i}>1 \\
      x^2_{i}, & \text{if}\ x_{i}<0 \\
      0, & \text{otherwise}
    \end{cases} \eqno{(1)} $$ 
$$ e = \frac{\sum_{ij}\overline{(\boldsymbol{E_{i}} - \boldsymbol{E_{j}})^2}}{M} \eqno{(2)} $$
$$ it = \frac{\sum_{i}\overline{\boldsymbol{IT_{i}}^2}}{N} \eqno{(3)} $$
$$ E65 = \frac{\overline{|\boldsymbol{E165}-\boldsymbol{v}|}+\overline{|\boldsymbol{E265}-\boldsymbol{v}|}}{2} \eqno{(4)} $$
$$ ST65 = \frac{\overline{|\boldsymbol{ST165}-\boldsymbol{f}|}+\overline{|\boldsymbol{ST265}-\boldsymbol{f}|}}{2} \eqno{(5)} $$

$d_{squared}$ is used to steer all the values of the array \textit{\(\boldsymbol{O}\)}, generated through the Real NVP inverse propagation, between 0 and 1. \textit{e} has the purpose of forcing \textit{M} pairs of voltages, \(\boldsymbol{E_{i}}\) and \(\boldsymbol{E_{j}}\), to be equal, with \textit{M} given by the number of pairs with a closed circuit breaker/relay between them (An example is the outputs of voltage sensors \(\boldsymbol{E_{i}} = \boldsymbol{E135}\) and \(\boldsymbol{E_{j}} = \boldsymbol{E140}\), separated by the circuit breaker CB136 in the high-left corner of Fig. \ref{fig:circuit_adapt}), \textit{it} is designed to set to 0 all \textit{N} currents, \(\boldsymbol{IT_{i}}\), measured in an open circuit, whereas $E65$ and $ST65$ steer a couple of voltages and frequencies to the values of 120.5V (\(\boldsymbol{v}\)) and 60Hz (\(\boldsymbol{f}\)), respectively, given their proximity to INV1 and INV2.

The vectors \(\boldsymbol{E}\), 
\(\boldsymbol{IT}\), 
\(\boldsymbol{E165}\), 
\(\boldsymbol{E265}\), 
\(\boldsymbol{ST165}\), 
\(\boldsymbol{ST265}\), 
\(\boldsymbol{v}\), and 
\(\boldsymbol{f}\) have a length equal to the time window given in input to the models, one of the hyper-parameters tuned, as described in the following section.

The physics-informed loss is equal to 
$$
    L_{phys\_inf} = d_{squared} + e + it + E65 + ST65 \eqno{(6)}
$$
The total loss is equal to:

$$
    L = L_{main} + \beta \cdot L_{phys\_inf} \eqno{(7)}
$$

Where $\beta$ is a weight learned through the Lagrangian dual framework \cite{lagr_paper}, which uses a Lagrangian dual approach to learn the best multipliers of a Lagrangian loss with a sub-gradient method.

$L_{main}$ is the main loss for each model, in the case of the GRU and Autoencoder models respectively the Mean Absolute Error (MAE) and the Mean Squared Error (MSE) were used, while for the Real NVP models $L_{main} = - L_{log\_prob}$, where $L_{log\_prob}$ is the log-likelihood of the current data.

The physics-informed loss was applied on the reconstructed array produced in output by the GRU and the Autoencoder, while, given that the Real NVP can be used to generate new data starting from a probability distribution, we applied this loss on the generated arrays. In each training step the Real NVP models can generate an arbitrary number of arrays, this number was selected through hyper-parameter tuning.

\subsection{Training \& Test}
\begin{figure}[t]
    \centerline{\includegraphics[width=0.93\linewidth]{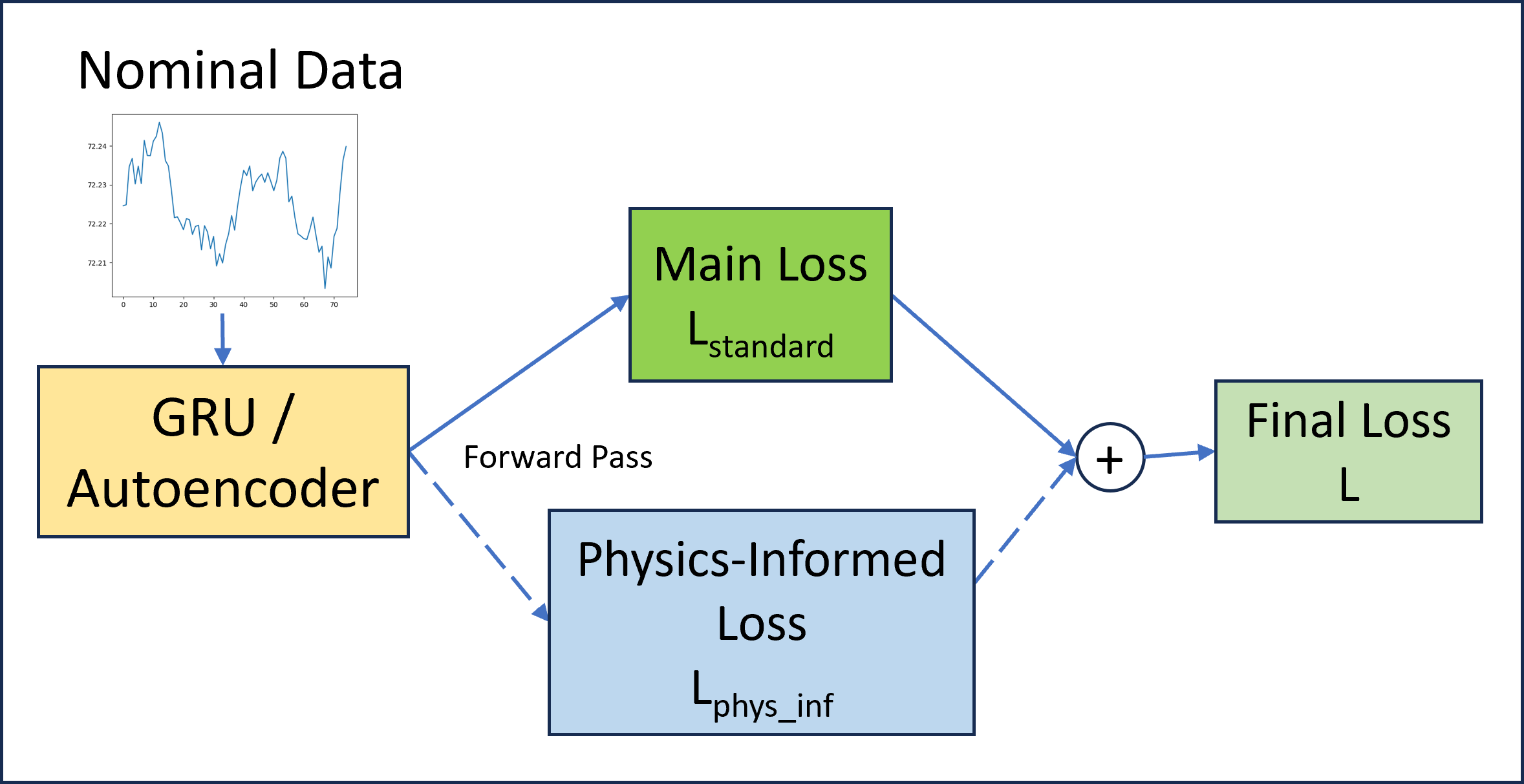}}
    \caption{Training method for GRU and Autoencoders. The models are tested with standard and physics-informed loss.}
    \label{fig:training_method_gru_auto}
\end{figure}

In Fig. \ref{fig:training_method_realnvp} and \ref{fig:training_method_gru_auto} we show how the models were trained. This was done in two different ways: all models received in input the nominal data, but while the GRU and the Autoencoder were trained to minimize the reconstruction error, the Real NVP was trained to increase the log-likelihood of the input data coming from the output distribution.

\begin{figure}[t]
    \centerline{\includegraphics[width=0.93\linewidth]{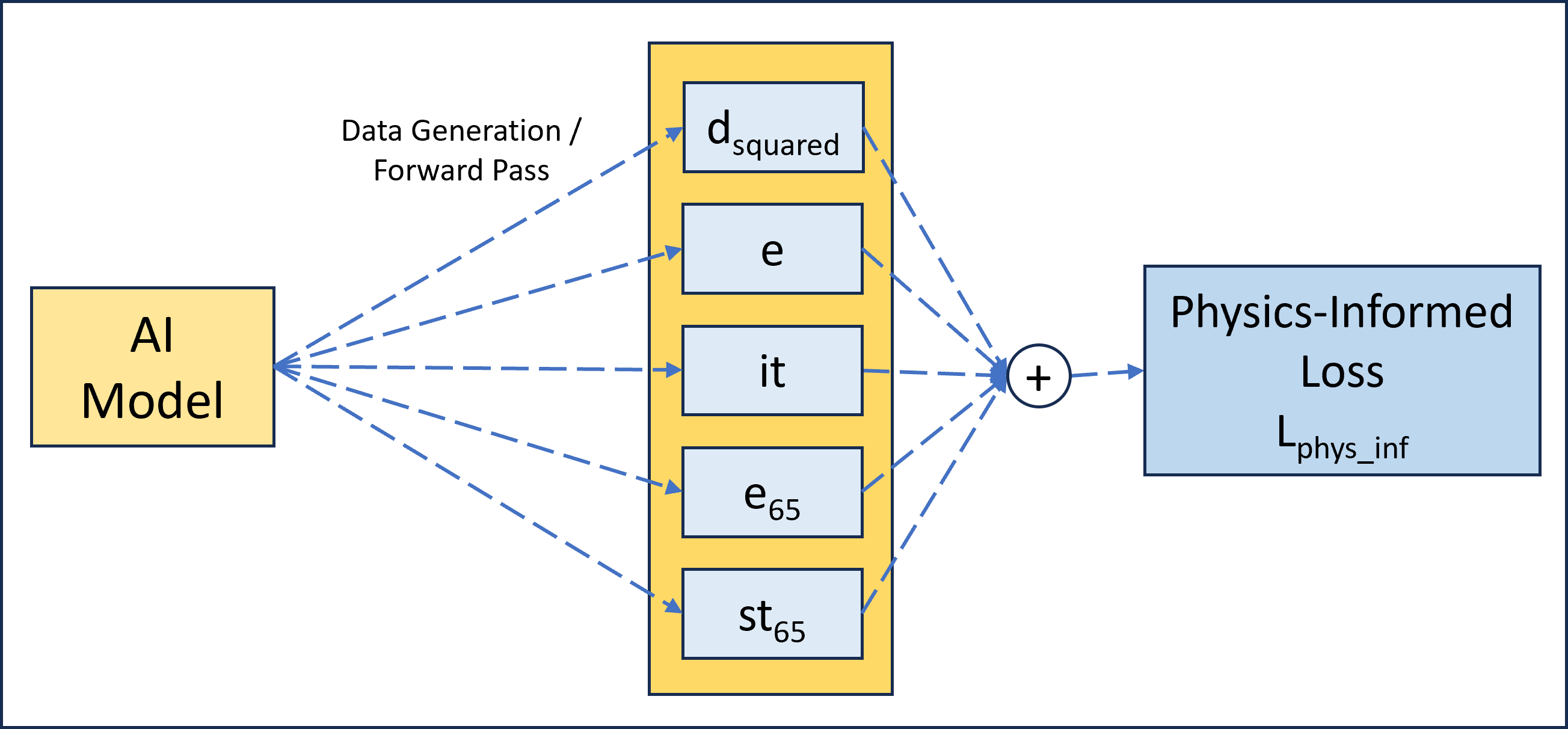}}
    \caption{Physics-informed loss composition (6).}
    \label{fig:ingr_pi}
    \vspace{-2mm}
\end{figure}

In Fig. \ref{fig:ingr_pi} is shown a graphical representation of how the above-mentioned Physics-Informed loss works.

During inference the output of the GRU and Autoencoder models was used to compute a reconstruction error, which can then be compared to a threshold to separate nominal and fault data, whereas the output of the Real NVP models, being a log-likelihood, can be directly used to separate nominal data from fault data, using a threshold.

\section{Experiments \& Results}
In the following, we first present how we conducted the experiments. Then, we report and analyze the results obtained.

\subsection{Experimental Setup}
Here we describe the dataset, the metrics used to evaluate the effectiveness of the proposed solution for the specific task, the libraries, and the computational resources used to perform the experiments.

\textbf{Dataset:} The dataset used for testing the model is ADAPT \cite{adapt_paper}, provided by NASA Ames Research Center. This dataset has been created with an EPS testbed to evaluate fault detection and isolation algorithms using controlled and repeatable fault injection scenarios. The dataset contains healthy and faulty conditions representative of the EPSs used in the aerospace sector.

\textbf{Metrics:} To evaluate the effectiveness of the Autoencoder, GRU, and Real NVP in detecting fault data in both dataset's splits and ensure their robustness and applicability, we consider the Area Under the ROC Curve (AUROC), from now on also abbreviated as AUC, False Positive Rate at 95\% true positive rate (FPR95), and F1-score.

AUC gives a comprehensive performance summary and is robust to imbalanced data, FPR95 gives a measure of the false positive rate at high specificity (95\%), this is particularly important in applications where false positives can have high costs. Finally, the F1-score is a typically used metric added here to ease comparisons. AUC and FPR95 let us analyze the performance of our models without selecting a threshold. While we use the threshold at FPR95 to compute the F1-score.

\textbf{Implementation Details:} We started with a comprehensive pre-processing phase, where the dataset was cleaned, scaled between 0 and 1, and two train-test splits were generated as described below.

A set of 10 random 70-30\% splits were created and a coarse grid search was launched using each one of them. Each grid search was performed using 30\% of the training data as validation set. Finally, the two more challenging splits that led to the worst AUC results on the test set were selected and used in all subsequent steps. This was done because an initial exploratory analysis and previous works \cite{mlp_adapt, cnn_adapt} showed that it was quite easy to obtain good results on ADAPT.

The training of all models comprised an initial phase of coarse grid search and a second phase of fine-grained grid search with 5 different random seeds.

During these phases, we optimized different hyper-parameters based on the model at hand, including the amount of past data used as input (i.e. the length of the input array), the number of layers, either fully-connected or GRU, and neurons of the NNs, the batch size, the number of coupling layers, and whether or not to use the physics-informed loss. All the ranges used for the hyper-parameters tuning are shown in Table \ref{tab:gridsearch}.

\begin{table}[t]
    \resizebox{\columnwidth}{!}{
    \begin{tabular}{cccccc}
        \toprule
        Model & \begin{tabular}{@{}c@{}}Past \\ Length\end{tabular} &\begin{tabular}{@{}c@{}}Coup.\\Layers\end{tabular} & Layers & Neurons & \begin{tabular}{@{}c@{}}Batch \\ Size\end{tabular} \\
        \midrule
        GRU - AE & \begin{tabular}{@{}c@{}}50, 30, \\ 10\end{tabular} & - & 2,4,6,8 &\begin{tabular}{@{}c@{}}512,256,\\128,64,32 \end{tabular} & \begin{tabular}{@{}c@{}}256,128, \\ 64,32\end{tabular} \\
        
        Real NVP & \begin{tabular}{@{}c@{}}50, 30, \\ 10\end{tabular} & 2,4,6 & 2,4,6,8 &\begin{tabular}{@{}c@{}}512,256,\\128,64,32 \end{tabular} & \begin{tabular}{@{}c@{}}256,128, \\ 64,32\end{tabular} \\
        \bottomrule
    \end{tabular}
    }
    \caption{Grid search combinations tested for each model. The coupling layers are used only for the Real NVP model.}
    \label{tab:gridsearch}
    \vspace{-2mm}
\end{table}

To perform these experiments we used custom models created with TensorFlow\footnote{https://www.tensorflow.org/}. Each experiment was run on an Intel Xeon Scalable Processors Gold 6130 with 16 cores and a frequency of 2.10 GHz.

\subsection{Results}
Table \ref{tab:results} summarizes the results obtained for both the dataset splits with the above-mentioned methods by averaging 5 different seeds during the grid search hyper-parameters optimization. The results reported in the table show the applications of both physics and not physics-informed loss. The PI-Real NVP works better in both splits. As shown in the table, the AUC of the ROC curve obtained from the PI-Real NVP is similar to the one of the Autoencoder; on the other hand, the FPR95 is lower, and the F1-score is higher in Real NVP. FPR95 is a critical score because it denotes how many false alarms the maintainer receives from the system. These scores confirm a reduction in the false positive detection with respect to other methods, making the PI-Real NVP more robust.

\begin{table}[t]
    \resizebox{\columnwidth}{!}{
    \begin{tabular}{cccccccccccccccccc}
        \toprule
        Split & Model & Past Length & AUC & FPR95 & F1-score \\
        \midrule
          & Real NVP & 50 & 0.94 & 0.142 & 0.95\\
          & PI-Real NVP & 50 & \textbf{0.95} & \textbf{0.101} & \textbf{0.95} \\
          & GRU & 50 & 0.91 & 0.301 & 0.92 \\
        1  & PI GRU & 50 & 0.93 & 0.249 & 0.93 \\
          & Autoencoder & 50 & 0.94 & 0.142 & 0.94 \\
          & PI Autoencoder & 50 & 0.95 & 0.108 & 0.95 \\
          & kPCA + MLP \cite{mlp_adapt} & - & 0.93 & 0.504 & 0.74 \\
        \midrule
          & Real NVP & 50 & 0.99 & 0.047 & 0.96 \\
          & PI-Real NVP & \textbf{10} & \textbf{0.99} & \textbf{0.005} & \textbf{0.98} \\
          & GRU & 50 & 0.96 & 0.146 & 0.94 \\
        2  & PI GRU & 50 & 0.96 & 0.135 & 0.94 \\
          & Autoencoder & 50 & 0.99 & 0.063 & 0.95 \\
          & PI Autoencoder & 50 & 0.99 & 0.041 & 0.96 \\
          & kPCA + MLP \cite{mlp_adapt} & - & 0.93 & 0.111 & 0.92 \\
        \bottomrule
    \end{tabular}
    }
    \caption{Results obtained for every split with each model averaged over 5 random seeds. For each result, we also show the amount of past data used. The PI-Real NVP model outperforms all the other models. Our model shows the best results in the second split, using less past data.}
    \label{tab:results}
\end{table}

Moreover, it can be noted that the models trained with the physics-informed loss achieve better performances with respect to the baselines, with the exception of the GRU models in the second split, which led to a slight improvement only for the FPR95. This shows that the physics-informed loss aids the fault detection process: it leads to a higher percentage of detected faults and reduces the number of false positives, thus providing more thrusthworty AI solutions.

Compared to previous AI studies on ADAPT, our models (and loss) achieve better performances with respect to the multivariate supervised approach \cite{mlp_adapt}, as shown in Table \ref{tab:results}, and to the univariate supervised approach \cite{cnn_adapt} reported in Table \ref{tab:results_cnn}.

\begin{table}[t]
    \resizebox{\columnwidth}{!}{
    \begin{tabular}{ccccccc}
        \toprule
        Split & Sensor & Model & Past Length & AUC & FPR95 & F1-score \\
        \midrule
        \multirow{ 2}{*}{1} & ST515 & PI-Real NVP & \textbf{50} & \textbf{0.92} & \textbf{0.08} & \textbf{0.81} \\
          & ST515 & ST + CNN \cite{cnn_adapt} & 120 & 0.77 & 0.93 & 0.15 \\
        \midrule
        \multirow{ 4}{*}{2} & E161 & PI-Real NVP & \textbf{10} & \textbf{0.99} & \textbf{0.00} & \textbf{0.98} \\
          & E161 & ST + CNN \cite{cnn_adapt} & 120 & 0.41 & 0.66 & 0.18 \\
          & ST515 & PI-Real NVP & 10 & 0.99 & 0.00 & 0.98 \\
          & ST515 & ST + CNN \cite{cnn_adapt} & 120 & 0.99 & 0.00 & 0.98 \\
        \bottomrule
    \end{tabular}
    }
    \caption{Results obtained when applying the method of \cite{cnn_adapt} on the columns of our splits with faulty data both in the train and test data. We tested our best models by removing from the test set all faults but those of the considered sensor. PI-Real NVP achieves superior or equivalent results.}
    \label{tab:results_cnn}
\end{table}

\cite{cnn_adapt} uses a CNN, preceded by a Stockwell Transform, to perform fault detection on a single sensor, i.e. a single column. It trains and tests each model only on a column of the dataset, thus dealing with an univariate distribution. While using a classifier for each sensor simplifies the data distribution and automatically covers also the fault isolation step, it requires to run inference on significantly more classifiers, thus leading to higher memory and computational costs. To compare our results to those of \cite{cnn_adapt}, for each split we: (1) selected the sensors with faulty measurements both in the train and in the test data; (2) trained and tested their method on each of those sensors data, performing a grid search for the parameters that we couldn't find in \cite{cnn_adapt}, i.e. kernel size \{1, 3, 5, 7\} and frequency index range \{100, 300, 500\}; (3) compared the results obtained with those of our best model (PI-Real NVP) evaluated by removing from the test set all anomalies except those linked to the considered sensor.

\cite{mlp_adapt} uses a MLP trained in a supervised fashion after having pre-processed the dataset with Kernel Principal Component Analysis (KPCA). Supervised approaches cannot be efficiently extended to datasets without, or with a very limited amount of labels as is often the case for satellite applications. In turn, our methods, and in particular Real NVP, which have been trained in a semi-supervised fashion, can be easily extended to incorporate a supervised loss. To train it we performed a grid search for the number of units of the fully-connected layers \{512, 256, 128, 64\} and for the batch size \{64, 32, 16\}, because we weren't able to find them in \cite{mlp_adapt}.

\begin{figure}[t]
    \centering
    \begin{subfigure}{0.94\linewidth}
        \includegraphics[width=\linewidth]{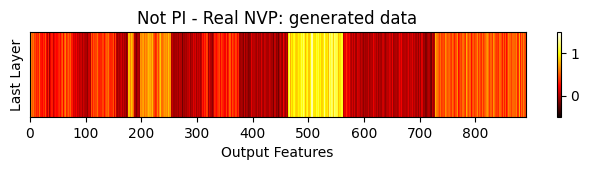}
    \end{subfigure}
    \begin{subfigure}{0.94\linewidth}
        \includegraphics[width=\linewidth]{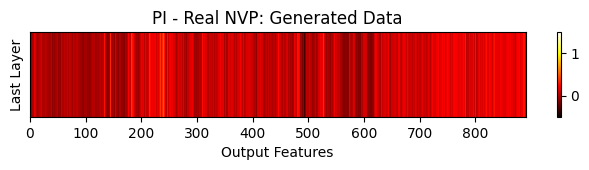}
    \end{subfigure}
    \caption{Data generated by the inverse propagation of the not PI-Real NVP, at the top, and the PI-Real NVP, at the bottom. On the x-axis are shown the output features for an experiment with a past window length of 10. These figures show that the PI loss successfully steered the Real NVP model towards the generation of data that doesn't violate the physics of the circuit, i.e. which falls in the range 0-1.}
    \label{fig:gen_data_last_layer}
\end{figure}

\textbf{Feature Maps Insights:} To provide a few insights on the effects of the physics-informed loss, we show the output of the last layer for the best performing model, i.e. the coupling layers of Real NVP, with and without PI loss. Fig. \ref{fig:gen_data_last_layer} shows the feature maps generated by the inverse propagation, i.e. when generating new data, of the non-PI and the PI models, respectively at the top and at the bottom of the figure. It can be noted that the features generated by the model trained without PI loss violate the physics of the circuit, as they are expected to be in the range 0-1, while those generated by the PI model fall in the given range. This demonstrates that the Real NVP successfully learned the \(d_{squared}\) sub-reward (1) and shows that the use of PI losses provides the means to improve the performance and explainability of the given AI model, resulting in a subsequent boost in confidence towards it.

\section{Conclusions \& Future Works}

This paper presents an application of Physics-Informed - Real NVP for Fault Detection. As explained in the previous sections, the experiments are performed on the ADAPT dataset by designing a physic-informed loss, which steers the model learning by providing physical insights into the data. This study shows that using PI-Real NVP leads to improved results on an EPS dataset built to be representative of satellite data coming from the same sub-system. Additionally, we showed that models trained with our physics-informed loss perform better with respect to those trained only with a standard loss. Given the high performances obtained in this dataset with a large set of AI models, we plan on testing the PI-Real NVP approach further by performing ablation studies on several parameters belonging both to the dataset and the models (e.g. number of input features, number of training data), and subsequently test it on new datasets representative both of EPS and of other satellites' sub-systems.





\section*{ACKNOWLEDGMENT}

This work has been developed with the contribution of the Politecnico di Torino Interdepartmental Centre for Service Robotics (PIC4SeR https://pic4ser.polito.it) and Argotec SRL, through the MUR – DM 117/2023 scholarship of Carlo Cena. Computational resources were provided by HPC@POLITO (http://hpc.polito.it).

\bibliographystyle{IEEEtran}
\bibliography{root}

\end{document}